\documentclass[conference]{IEEEtran}
\IEEEoverridecommandlockouts
\usepackage{amsmath,amssymb,amsfonts,epsfig}
\usepackage{cite}
\usepackage{bm}
\usepackage{algorithmic}
\usepackage{graphicx}
\usepackage{textcomp}
\usepackage{xcolor}
\usepackage{multirow}
\usepackage{booktabs}
\usepackage{url}
\usepackage{subcaption}
\usepackage{threeparttable}
\usepackage{pifont}
\newcommand{\cmark}{\ding{51}}%
\def\mathbi#1{\textbf{\em #1}}
\def\BibTeX{{\rm B\kern-.05em{\sc i\kern-.025em b}\kern-.08em
    T\kern-.1667em\lower.7ex\hbox{E}\kern-.125emX}}
\begin{document}

\title{SimCast: Enhancing Precipitation Nowcasting with Short-to-Long Term Knowledge Distillation}

\author{
\IEEEauthorblockN{Yifang Yin\IEEEauthorrefmark{1}, Shengkai Chen\IEEEauthorrefmark{1}, Yiyao Li\IEEEauthorrefmark{2}, Lu Wang\IEEEauthorrefmark{1}, Ruibing Jin\IEEEauthorrefmark{3}, Wei Cui\IEEEauthorrefmark{1}, Shili Xiang\IEEEauthorrefmark{1}}
  \IEEEauthorblockA{Institute for Infocomm Research (I$^2$R), A*STAR, Singapore\\
  \IEEEauthorrefmark{1}\{yin\_yifang, chen\_shengkai, wang\_lu, cui\_wei, sxiang\}@i2r.a-star.edu.sg} \IEEEauthorrefmark{2}lyyleo2001@gmail.com, \IEEEauthorrefmark{3}ruibing\_jin@outlook.com
}

\maketitle

\begin{abstract}
Precipitation nowcasting predicts future radar sequences based on current observations, which is a highly challenging task driven by the inherent complexity of the Earth system. Accurate nowcasting is of utmost importance for addressing various societal needs, including disaster management, agriculture, transportation, and energy optimization. As a complementary to existing non-autoregressive nowcasting approaches, we investigate the impact of prediction horizons on nowcasting models and propose SimCast, a novel training pipeline featuring a short-to-long term knowledge distillation technique coupled with a weighted MSE loss to prioritize heavy rainfall regions. Improved nowcasting predictions can be obtained without introducing additional overhead during inference. As SimCast generates deterministic predictions, we further integrate it into a diffusion-based framework named CasCast, leveraging the strengths from probabilistic models to overcome limitations such as blurriness and distribution shift in deterministic outputs. Extensive experimental results on three benchmark datasets validate the effectiveness of the proposed framework, achieving mean CSI scores of 0.452 on SEVIR, 0.474 on HKO-7, and 0.361 on MeteoNet, which outperforms existing approaches by a significant margin. 

\end{abstract}
\section{Introduction}
\label{sec:intro}
Precipitation nowcasting aims at generating forecasts of rainfall intensity for up to 2 hours in local areas, typically using radar echo maps as the primary data source~\cite{veillette2020sevir}. This is a highly challenging task due to the inherent complexity of the Earth system, where accurate predictions are often hampered by unpredictable weather fluctuations. Nevertheless, achieving skillful precipitation nowcasting is critically important for addressing societal needs and mitigating the impacts of adverse weather events. The demand for robust nowcasting spans diverse applications, including disaster management, agricultural planning, transportation safety, energy optimization, \emph{etc.} Thus, this problem has drawn increasing attention from the research community in recent years~\cite{gao2022earthformer,gong2024cascast,yu2024diffcast}.

With the growing availability of large-scale radar echo benchmark datasets such as HKO-7~\cite{shi2017deep} and SEVIR~\cite{veillette2020sevir}, data-driven deep learning techniques have emerged as the leading approach in recent advancements in precipitation nowcasting. Deterministic models including CNN-based~\cite{veillette2020sevir,tan2025simvpv2}, RNN-based~\cite{wang2017predrnn,wang2022predrnn}, and Transformer-based~\cite{gao2022earthformer} deep neural networks have proven effective in predicting the development and dissipation of precipitation, addressing the limitations of traditional extrapolation-based methods. Methods such as PhyDNet~\cite{guen2020disentangling} and NowcastNet~\cite{zhang2023skilful} explicitly model physical dynamics of weather systems, encouraging predictions to align with the underlying physical constraints. In recent years, probabilistic models including GAN-based~\cite{zhang2023skilful} and diffusion-based~\cite{gao2024prediff} approaches have been proposed to mitigate the blurriness commonly observed in deterministic outputs, but encounter challenges in achieving overall prediction accuracy across different scales of weather events~\cite{gong2024cascast}. To solve this issue, a recent direction involves training a probabilistic model conditioned on deterministic predictions to enhance its perceptual quality~\cite{gao2024prediff,yoon2025probabilistic}. Studies have shown that high-quality deterministic predictions generally lead to more accurate nowcasting results generated by probabilistic models.

While it is common to model the physical constraints between consecutive frames using an autoregressive mechanism (\emph{e.g.}, EvolutionNet~\cite{zhang2023skilful}), most deep learning-based nowcasting models are non-autoregressive methods that generate all predictions in parallel. The influence of the prediction horizon remains underexplored in precipitation nowcasting research. To bridge this gap, we investigate the varying behaviors of nowcasting models when trained using different prediction horizons, and observe that prediction horizon plays an important role in enabling the model to capture temporal dynamics within different ranges. This insight motivates us to first train a short-term nowcasting model to accurately capture near-future precipitation patterns and then transfer its knowledge to a longer-term nowcasting model. Specifically, as the training of nowcasting models requires long and diverse radar sequences, which are often constrained by memory and storage limitations, we propose leveraging the short-term nowcasting model to augment training sequences by appending synthetic radar images to their endpoints. A random sub-sequence sampling strategy is employed within each batch to enhance the diversity of training samples and mitigate overfitting.  
Combined with the weighted MSE loss, our short-to-long term knowledge distillation approach enables the long-term nowcasting model to learn from both ground-truth radar data and short-term forecasts, with a particular emphasis on heavy rainfall areas. 

We compared our method to both deterministic and probabilistic nowcasting models on three benchmark radar datasets, namely SEVIR, HKO-7, and MeteoNet, where the new state-of-the-art nowcasting results have been obtained, especially
in high-intensity rainfall areas. 
Importantly, during inference, the long-term nowcasting model in our framework generates all future frames in a non-autoregressive manner, ensuring that our method introduces no additional computational overhead when deployed in practical applications.
Furthermore, we integrate our method with CasCast~\cite{gong2024cascast} to cope with the common issue of blurriness in deterministic predictions. Experimental results demonstrate that using enhanced deterministic predictions as conditional inputs for diffusion models leads to improved nowcasting performance in both accuracy and perceptual quality.

\section{Related Work}
Deep learning based precipitation nowcasting methods can be classified into the following three categories: deterministic models, probabilistic models, and their fusion. Popular deterministic models for precipitation nowcasting include ConvLSTM~\cite{shi2015convolutional}, PredRNN~\cite{wang2017predrnn,wang2022predrnn}, PhyDNet~\cite{guen2020disentangling}, SimVP~\cite{gao2022simvp}, and EarthFormer~\cite{gao2022earthformer}. For example, PhyDNet~\cite{guen2020disentangling} is a two-branch deep model, which disentangles physical dynamics from residual information. Physical dynamics is modeled by a newly proposed recurrent physical cell termed PhyCell, and residual information is modeled by a data-driven ConvLSTM~\cite{shi2015convolutional} model. SimVP~\cite{gao2022simvp} is an end-to-end CNN model trained by a simple MSE loss, obtaining state-of-the-art performance on multiple video prediction and radar-based nowcasting benchmark datasets. Earthformer~\cite{gao2022earthformer} is a Transformer model designed for Earth system forecasting. A generic and flexible space-time attention block, named Cuboid Attention, is proposed to support efficient 3D attentions. Deterministic models focus on capturing the overall motion by providing a single-value prediction of the future state. However, this approach often results in blurry predictions and a lack of fine-grained details, particularly for longer lead times.

To generate high-fidelity prediction results, probabilistic models have recently been proposed for radar-based nowcasting~\cite{ravuri2021skilful,yan2021videogpt,rombach2022high,gao2024prediff}. PreDiff~\cite{gao2024prediff} represents one of the early efforts to achieve probabilistic forecasts using a conditional latent diffusion model. DGMR~\cite{ravuri2021skilful}, VideoGPT~\cite{yan2021videogpt}, and LDM~\cite{rombach2022high} addressed the issue of blurry predictions, but they often suffer from lower CSI scores. Recent research has increasingly focused on integrating the strengths of deterministic and probabilistic models~\cite{zhang2023skilful,yoon2025probabilistic,gong2024cascast,yu2024diffcast}. DiffCast~\cite{yu2024diffcast} proposed to model the precipitation evolution from the perspective of global deterministic motion and local stochastic variations with residual mechanism. NowcastNet~\cite{zhang2023skilful} models the physical evolution of weather systems using a deterministic model, which is then employed to train a conditional GAN with adversarial loss. CasCast~\cite{gong2024cascast} presented a cascaded framework consisting of deterministic and probabilistic components, trained sequentially in a cascaded manner. It offers greater flexibility by demonstrating that improved deterministic conditions can lead to more accurate final nowcasting results.

\section{Our Approach}
\begin{figure}[t!]
    \centering
    \begin{subfigure}[t]{\linewidth}
        \centering
    \includegraphics[width=0.8\linewidth]{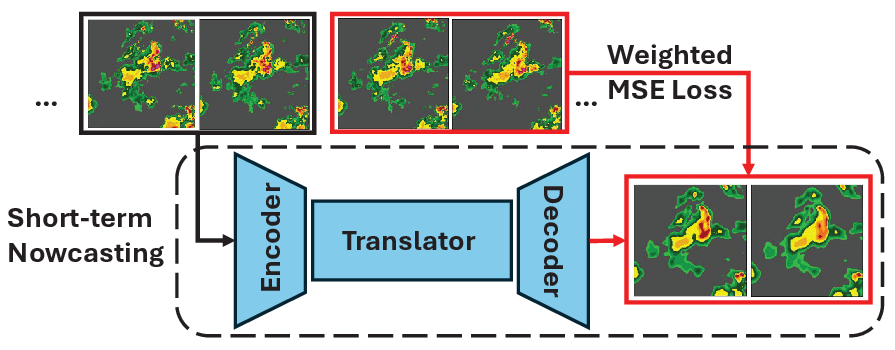}
        \caption{Training a short-term precipitation nowcasting model.}
    \end{subfigure}%
    \vspace{1mm}
    \begin{subfigure}[t]{\linewidth}
        \centering
        \includegraphics[width=1\linewidth]{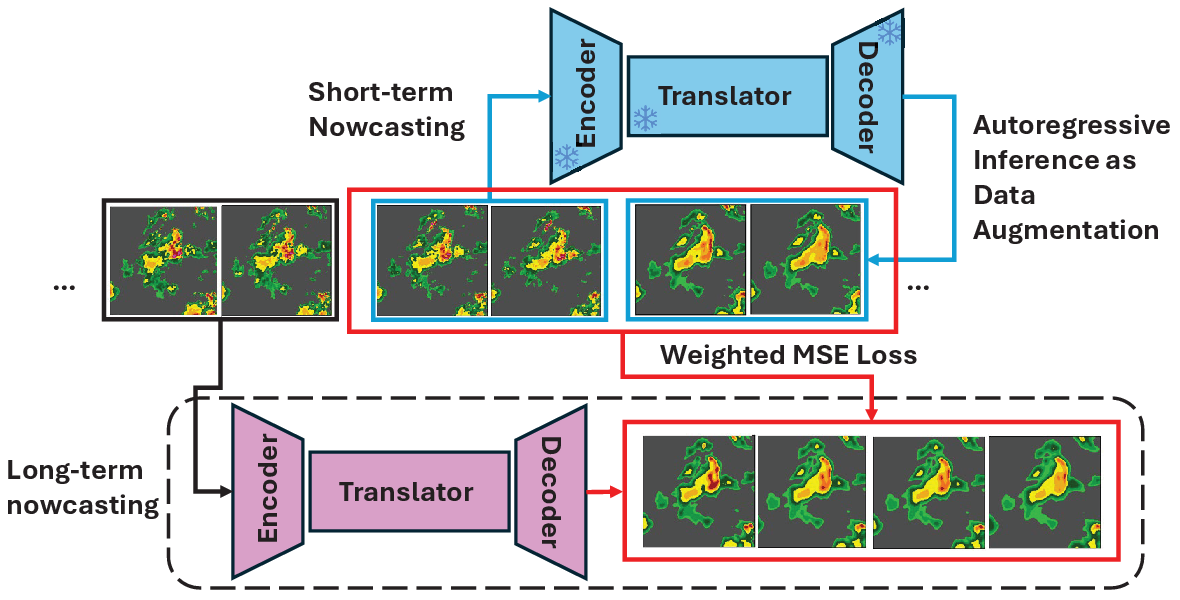}
        \caption{Autoregressively apply the short-term nowcasting model to generate synthetic radar data, appending them to the end of each training sample. Subsequently, train a long-term precipitation nowcasting model using the augmented training dataset.}
    \end{subfigure}
    \caption{Illustration of our proposed short-term to long-term knowledge distillation for enhanced precipitation nowcasting.}
\label{fig:overview}
\vspace{-5mm}
\end{figure}

Our proposed method consists of two major stages. As illustrated in Fig.~\ref{fig:overview}, SimCast begins by training a short-term precipitation nowcasting model, focusing on capturing robust patterns of short-term precipitation dynamics. In the subsequent stage, we introduce a novel short-term to long-term knowledge distillation method, enabling the transfer of knowledge from the short-term model to develop a more accurate long-term precipitation nowcasting model.

\subsection{Precipitation Nowcasting}
Precipitation nowcasting is commonly formalized as a spatiotemporal forecasting problem~\cite{veillette2020sevir}. Given radar observations $\mathbi{X}_{t,T}=\{x_i\}_{t-T+1}^{t}$ from the past $T$ timestamps, the goal is to predict the future radar sequence $\mathbi{Y}_{t,T^\prime}=\{x_i\}_{t+1}^{t+T^\prime}$, where $t$ denotes the current time and $T^\prime$ is the prediction horizon. $x_i \in \mathbb{R}^{C,H,W}$ represents a radar image, with the channel dimension $C=1$, indicating the intensity of
radar echoes.

Following the above problem formulation, we denote the prediction horizons for the short-term and long-term precipitation nowcasting in our framework as $T_s^\prime$ and $T_l^\prime$, respectively. Notably, both the short-term and long-term nowcasting models share the same underlying architecture. In the rest of this section, we first present the technical details of our nowcasting model, followed by an introduction to our newly proposed short-term to long-term knowledge distillation approach for enhanced radar-based nowcasting.

\subsubsection{Model Architecture}
We adopt SimVP~\cite{gao2022simvp} as our nowcasting model due to its simplicity and superior performance over various spatiotemporal forecasting problems. It consists of an \emph{encoder} that extracts spatial features, a \emph{translator} that learns temporal evolutions, and a \emph{decoder} that reconstructs future radar frames. All three components are constructed using CNNs. While the encoder and decoder convolute $C$ channels on $(H, W)$, the translator convolutes $T \times C$ channels on $(H, W)$ to capture temporal dynamics. Efforts have been made on improving the spatiotemporal predictive learning capability of the translator, including approaches such as Inception-Unet~\cite{szegedy2015going}, gSTA~\cite{tan2025simvpv2}, and TAU~\cite{tan2023temporal}. In this work, we adopt the Inception-Unet architecture, which, based on our empirical studies, exhibits robust performance in radar-based nowcasting tasks.

\subsubsection{Optimization}
For radar-based nowcasting, predicting high-intensity rainfall pixels is particularly critical yet poses greater challenges in achieving high accuracy, especially for longer lead times. Existing studies often assign pixel-wise weights, which are calculated based on rainfall intensities, to address data imbalance and prioritize regions with high-intensity rainfall~\cite{ravuri2021skilful,zhang2023skilful}. However, in practice, the importance of pixels may not solely determined by rainfall intensity, and explicitly modeling fine-grained pixel-wise weights may lead to suboptimal perceptual loss. Based on the observations, we optimize our model based on a simpler yet effective weighted MSE loss as,
\begin{equation}
    \mathcal{L} = \sum_{T,H,W}\omega(\mathbi{Y}_{t,T^\prime})\cdot (f_\theta(\mathbi{X}_{t,T}) - \mathbi{Y}_{t,T^\prime})^2
\end{equation}
where $f_\theta$ denotes our nowcasting model with learnable parameters $\theta$. We pre-define a rainfall intensity threshold $\tau$, and compute the pixel-wise weight $\omega$ as,
\begin{equation}
    \omega(x) = \left \{
    \begin{array}{rcl}
 1 ~~~& \mbox{if}~x\leq \tau\\  
 w_{max} & \mbox{otherwise} \\
\end{array}
\right.
\label{eq:pixelwise_weight}
\end{equation}

One practical strategy is to set $\tau$ as the threshold corresponding to the highest rainfall category to be predicted in specific applications, \emph{e.g.}, 219 for SEVIR benchmark dataset. Based on our empirical studies, we observe that this strategy performs robustly across multiple datasets with $w_{max}=10$. 

\subsubsection{Inference} Once a model is trained, it can be applied autoregressively multiple times to generate predictions of arbitrary length. This strategy is commonly adopted, particularly by computational intensive models such as diffusion models, where the prediction horizon is constrained by computational resources~\cite{hoppe2022diffusion}. However, the selection of the prediction horizon plays a critical role in shaping the temporal dynamics captured by the model. This motivates us to propose a new short-term to long-term knowledge distillation approach to integrate the strengths of nowcasting models that capture varying temporal patterns.

\subsection{Short-to-Long Term Knowledge Distillation}
\begin{figure}[t!]
    \centering
    \includegraphics[width=0.95\linewidth]{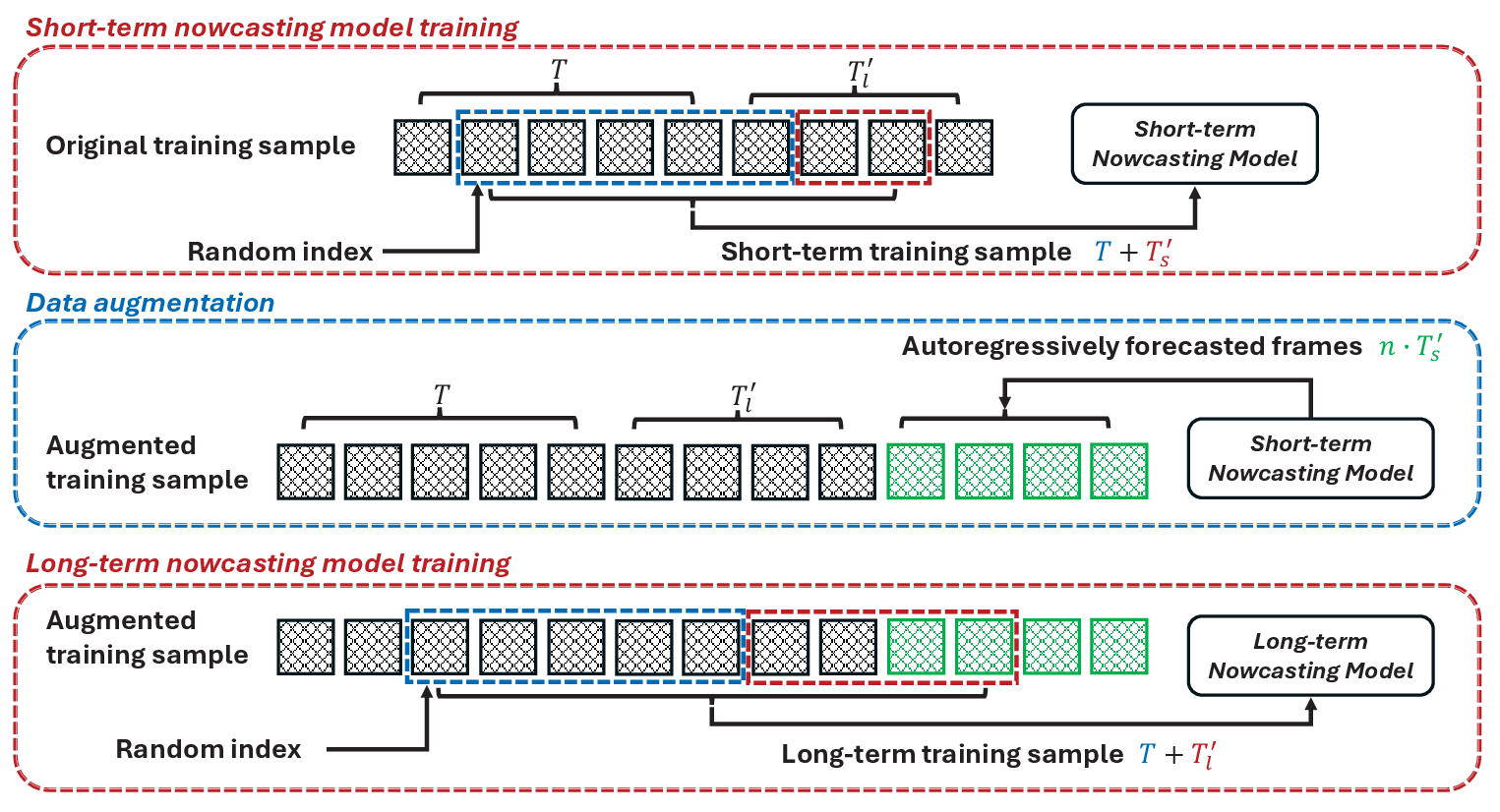}
    \caption{The proposed three-step training strategy with data augmentation and knowledge distillation.}
    \label{fig:method}
\vspace{-5mm}
\end{figure}

Recall that $T$ denotes the number of past radar observations, $T_s^\prime$ and $T_l^\prime$ represent the number of future radar images to be predicted by the short-term and long-term nowcasting models, respectively. Once the short-term nowcasting model is trained, we propose utilizing it as a data augmentation approach to extend the length of training samples by appending forecasted synthetic radar images to their end. As shown in Fig.~\ref{fig:method}, our approach is different from traditional knowledge distillation techniques, offering the following key advantages. 

\emph{Firstly}, the augmented training samples are enriched by a greater number of radar frames, which enables the random sampling of sub-sequences within each batch to train nowcasting models. This significantly improves the diversity of training samples, which is critical for spatiotemporal predictive learning problems. \emph{Secondly}, as shown at the bottom of Fig.~\ref{fig:method}, with the random index increasing from 0, the long-term training samples progressively include more synthetic radar images generated by our short-term nowcasting models. Therefore, a simple MSE loss allows our long-term nowcasting model to learn from both ground-truth radar data and short-term forecasts. This aligns with knowledge distillation~\cite{yin2021enhanced} to harvest its benefits. \emph{Lastly}, randomly sampling of sub-sequences from training samples is also applicable to short-term model training (see top of Fig.~\ref{fig:method}), owing to the reduced look-ahead time. This enables the generation of more robust short-term forecasts to facilitate effective knowledge transfer in the subsequent step.

The final nowcasting model in our framework is non-autoregressive, which produces $T_l^\prime$ frames in parallel, without relying on previously generated samples. Thus, our proposed method does not introduce any overhead during inference, while achieving significant performance improvements.

\section{Evaluation}
\subsection{Experimental Setup}
\noindent \textbf{Datasets.} We conducted experiments using three benchmark datasets, namely SEVIR~\cite{veillette2020sevir}, HKO-7~\cite{shi2017deep}, and MeteoNet~\cite{larvor2021meteonet}. We preprocessed HKO-7 and MeteoNet following CasCast~\cite{gong2024cascast}, and the statistics of the datasets are presented in Table~\ref{tab:dataset}.
\vspace{-2mm}
\begin{table}[h]
\centering
\resizebox{\linewidth}{!}{
\begin{threeparttable}[b]
    \caption{Statistics of the three radar echo benchmark datasets.}
\begin{tabular}{c|cccccccc}
\toprule
Dataset & $N_{train}$ & $N_{val}$ & $N_{test}$ & Resolution & Size & Interval & $L_{in}$ & $L_{out}$ \\
\midrule
SEVIR & 35,718 & 9,060 & 12,159 & 1 km & 384 & 5 min & 13 & 12 \\
HKO-7 & 8,772 & 492 & 1,152 & 2 km & 480 & 6 min & 10 & 10 \\
MeteoNet & 6,978 & 2,234 & 994 & 0.01$^{\circ}$ &400 & 5 min & 12 & 12\\
\bottomrule
\end{tabular}
\label{tab:dataset}
\end{threeparttable}}
\vspace{-2mm}
\end{table}

\begin{table*}[thbp]
\centering
\resizebox{0.9\textwidth}{!}{
\begin{threeparttable}[b]
    \caption{Comparison with the state-of-the-art radar-based nowcasting approaches on SEVIR dataset. *Following CasCast~\cite{gong2024cascast}, EarthFormer is evaluated using the official checkpoint.}
\begin{tabular}{c|c|c|c|ccc|ccc|ccc}
\toprule
\multirow{2}{*}{Method}&\multirow{2}{*}{CRPS} &\multirow{2}{*}{SSIM} &\multirow{2}{*}{HSS}&\multicolumn{3}{c|}{CSI-M} & \multicolumn{3}{c|}{CSI-181} & \multicolumn{3}{c}{CSI-219} \\ 

&&&&POOL1&POOL4&POOL16&POOL1&POOL4&POOL16&POOL1&POOL4&POOL16\\
\midrule
\midrule
ConvLSTM &0.0264& 0.7749& 0.5232& 0.4102 &0.4163& 0.4475 &0.2453 &0.2525 &0.2977 &0.1322 &0.1380 &0.1734\\
PredRNN&0.0271 &0.7497 &0.5192& 0.4045& 0.4161 &0.4623& 0.2416 &0.2567 &0.3214 &0.1331& 0.1447& 0.1909\\
PhyDNet &0.0253 &0.7649 &0.5311 &0.4198 &0.4226 &0.4410 &0.2526& 0.2532 &0.2782 &0.1362 &0.1359 &0.1526\\
SimVP& 0.0259 &\textbf{0.7772} &0.5280 &0.4153 &0.4226 &0.4530 &0.2532 &0.2604 &0.3000 &0.1338 &0.1394 &0.1685\\
EarthFormer*& \textbf{0.0251} &0.7756 &0.5411 &0.4310 &0.4319 &0.4351 &0.2622 &0.2542 &0.2562& 0.1448& 0.1409& 0.1481\\
SimCast, ours &0.0270 & 0.7252&\textbf{0.5834} & \textbf{0.4521} &\textbf{0.4750} & \textbf{0.4968}&\textbf{0.3099} &\textbf{0.3343}& \textbf{0.3571} & \textbf{0.2007}& \textbf{0.2517}& \textbf{0.3268}\\
\midrule
LDM& 0.0208 &0.7495 &0.4386 &0.3465 &0.3442 &0.3520 &0.1470 &0.1391 &0.1432 &0.0671 &0.0655 &0.0717\\
PreDiff & \textbf{0.0202} &0.7648& 0.4914& 0.3875 &0.3918 &0.4157 &0.2076 &0.2069& 0.2264 &0.1032& 0.1051 &0.1213\\
NowcastNet & 0.0283& 0.5696& 0.5365& 0.4152& 0.4452& 0.5024 &0.2495 &0.2935 &0.3725 &0.1422 &0.1874& 0.2700\\
CasCast(EarthFormer)&\textbf{0.0202}& \textbf{0.7797} &0.5602& 0.4401 &0.4640 &0.5225 &0.2879 &0.3179 &0.3900 &0.1851 &0.2127 &0.2841\\
CasCast(SimCast), ours & 0.0259& 0.7620& \textbf{0.5771}& \textbf{0.4467}& \textbf{0.4811} & \textbf{0.5501}& \textbf{0.3049} & \textbf{0.3470}& \textbf{0.4252}& \textbf{0.1856} & \textbf{0.2353} & \textbf{0.3387}\\
\bottomrule
\end{tabular}
\label{tab:eval_sevir}
\end{threeparttable}}
\vspace{-1mm}
\end{table*}

\begin{table*}[thbp]
\centering
\resizebox{0.9\textwidth}{!}{
\begin{threeparttable}[b]
    \caption{Comparison with the state-of-the-art radar-based nowcasting approaches on HKO-7 and MeteoNet datasets.}
\begin{tabular}{c|ccc|ccc|ccc|ccc}
\toprule
\multirow{3}{*}{Method}& \multicolumn{6}{c|}{HKO-7}&\multicolumn{6}{c}{MeteoNet}\\
\cline{2-13}
&\multicolumn{3}{c|}{CSI-M} & \multicolumn{3}{c|}{CSI-185} & \multicolumn{3}{c|}{CSI-M} & \multicolumn{3}{c}{CSI-47} \\ 
&POOL1&POOL4&POOL16&POOL1&POOL4&POOL16&POOL1&POOL4&POOL16&POOL1&POOL4&POOL16\\
\midrule
\midrule
ConvLSTM &0.4000 &0.4084 &0.4280 &0.1569 &0.1843 &0.2472&0.3008& 0.3050 &0.3465 &0.0982 &0.1091 &0.1588\\
PredRNN &0.3996 &0.4146 &0.4398 &0.1633& 0.1981 &0.2634 &0.2914 &0.3003 &0.3402 &0.0823 &0.0990 &0.1462\\
PhyDNet &0.4213 &0.4121 &0.3846 &0.1807 &0.1768 &0.1913 &0.3120 &0.3124 &0.3356 &0.1106 &0.1157 &0.1482\\
SimVP &0.4236 &0.4195 &0.4134 &0.1881 &0.1953 &0.2233 &0.3017 &0.3143 &0.3577 &0.0997 &0.1134 &0.1599\\
EarthFormer &0.4096 &0.4003 &0.3950 &0.1729 &0.1731 &0.1935 &0.2831 &0.2855 &0.3154 &0.0787 &0.0872 &0.1208\\
\midrule
LDM&0.3045 &0.2738 &0.2764 &0.0517 &0.0605 &0.0928 &0.2131 &0.2191 &0.2369 &0.0359 &0.0407 &0.0552 \\
PreDiff &0.3221 &0.3152 &0.3046 &0.0788 &0.0852& 0.1113&0.2546 &0.2668 &0.2935 &0.0490 &0.0594 &0.0867\\
NowcastNet &0.4234 &0.4518 &\underline{0.4724} &0.2025& 0.2607 &0.3601 &0.2955 &0.3232 &\underline{0.3734} &\underline{0.1236} &0.1521 &0.2115\\
CasCast(EarthFormer)& \underline{0.4267} &\underline{0.4608} &\textbf{0.4938} &\underline{0.2158}& \underline{0.2772}& \underline{0.3653}&\underline{0.3156} &\textbf{0.3650} &\textbf{0.4420} &0.1204 &\underline{0.1563}& \underline{0.2357}\\
\midrule
SimCast, ours &\textbf{0.4740}&\textbf{0.4779}&0.4390&\textbf{0.2829}&\textbf{0.3517}&\textbf{0.3744}&\textbf{0.3610}&\underline{0.3564} & 0.3506&\textbf{0.1922}&\textbf{0.2173} &\textbf{0.2548}\\
\bottomrule
\end{tabular}
\label{tab:eval_others}
\end{threeparttable}}
\vspace{-4mm}
\end{table*}

\noindent \textbf{Evaluation Metrics.} We report Critical Success Index (CSI), Structural Similarity Index (SSIM), Heidke Skill Score (HSS), and Continuous Ranked Probability Score (CRPS) to assess the performance of precipitation nowcasting models. Following previous work, we also report CSI at $4 \times 4$ and $16 \times 16$ max pool scales to evaluate the model's capability in capturing local precipitation patterns and predicting regional extreme precipitation events. CRPS measures the effectiveness of uncertainty modeling in probabilistic predictions and simplifies to the MAE when applied to deterministic predictions.

\noindent \textbf{Implementation Details.} We followed the dataset configuration and set the prediction horizon $T_l^\prime$ to 12, 10, and 12 for SEVIR, HKO-7, and MeteoNet, respectively. The short-term prediction horizon was set to $T_s^\prime=6, 5, 6$, which is half of $T_l^\prime$ by default. The short-term nowcasting model was applied autoregressively twice to augment the original samples with $T_l^\prime$ frames appended to the end of each training sequence. We implemented our nowcasting model with 2 spatial blocks in both the encoder and decoder and 4 temporal blocks in the translator. The threshold $\tau$ for computing pixel-wise weight was set to 219, 185, and 47, the highest rainfall category in SEVIR, HKO-7, and MeteoNet, respectively. $w_{max}$ was set to 10 by default. We trained our model using the Adam optimizer with a learning rate of 0.005 and a batch size of 8. Training was conducted for a maximum of 100 epochs. For system efficiency, our framework uses the short-term nowcasting model as the teacher model for knowledge distillation, introducing a training overhead of around 0.03 s per sample on a single A40 GPU, similar to a standard teacher-student framework. 

\subsection{Comparison to the State-of-the-arts}
We compared SimCast with the following 9 state-of-the-art nowcasting models: ConvLSTM\cite{shi2015convolutional}, PredRNN\cite{wang2017predrnn}, PhyDNet\cite{guen2020disentangling}, SimVP\cite{gao2022simvp}, EarthFormer\cite{gao2022earthformer}, LDM\cite{rombach2022high}, PreDiff\cite{gao2024prediff}, NowcastNet\cite{zhang2023skilful}, and CasCast\cite{gong2024cascast} on SEVIR, HKO-7, and MeteoNet. The results are presented in Tables~\ref{tab:eval_sevir} and~\ref{tab:eval_others}.

We first compared our proposed SimCast with existing deterministic models, including ConvLSTM, PredRNN, PhyDNet, SimVP, and EarthFormer, on the SEVIR dataset. Following previous work, we compute CSI at thresholds 16, 74, 133, 160, 181, and 219. CSI-M refers to the mean CSI over all thresholds. As can be seen, SimCast obtained the best HSS, CSI-M, CSI-181, and CSI-219 at all pooling scales. The results indicate that our proposed weighted MSE loss and short-to-long term knowledge distillation can effectively improve the overall prediction accuracy, especially for heavy rainfall regions. However, employing a weighted MSE loss may cause the model's output to deviate from true radar distribution, leading to a reduction in the SSIM, a perceptual metric that quantifies image quality degradation. 

Fortunately, the bias caused by weighted MSE loss can be effectively corrected by combining the strengths of probabilistic models. To evaluate, we integrate the proposed SimCast with CasCast by leveraging SimCast's predictions as conditional inputs to train a diffusion-based CasFormer~\cite{gong2024cascast}, refining the outputs to improve visual quality. Then we compared the integrated CasCast(SimCast) method with four probabilistic models including LDM, PreDiff, NowcastNet, and CasCast(EarthFormer). To be consistent with previous work~\cite{gong2024cascast}, each probabilistic model was run multiple times, and the results were compared using an ensemble of 10 runs. LDM and PreDiff are diffusion-based models conditioned on past radar observations $\mathbi{X}_{t,T}$. NowcastNet and CasCast(EarthFormer) are additionally conditioned on the outputs of separate deterministic branches, namely EvolutionNet and EarthFormer, respectively. Thus, improved prediction accuracy has been achieved. By substituting EarthFormer with the proposed SimCast as the deterministic branch in the CasCast framework, CasCast(SimCast) achieved the highest HSS and CSI scores in all cases. Moreover, the perceptual metric SSIM was significantly improved from 0.7252 to 0.7620.

For HKO-7 and MeteoNet datasets, a similar pattern can be observed in Table~\ref{tab:eval_others}. SimCast significantly improved the prediction accuracy for heavy rainfall, achieving a CSI-185 of 0.2829 on HKO-7 and a CSI-47 of 0.1922 on MeteoNet that substantially outperformed all deterministic and probabilistic models. The mean CSI (CSI-M) is calculated by averaging the CSI scores over a predefined set of thresholds: [84, 118, 141, 158, 185] for HKO-7 and [19, 28, 35, 40, 47] for MeteoNet. When converting the pixel values to rain rates (mm/h), both set of pixel thresholds correspond to the same rain rate thresholds, which are [0.5, 2, 5, 10, 30]. 

\begin{table}[tbp]
\small
\centering
\resizebox{0.95\columnwidth}{!}{
\begin{threeparttable}[b]
    \caption{Model design ablation studies using the SEVIR dataset.}
\begin{tabular}{c |c c |c c}
\toprule
Model &Weighted MSE& Short-to-long KD & CSI-M &Gain \\ 
\midrule
SimCast\textsubscript{12} & & - &0.4166&-\\
SimCast\textsubscript{12} &\cmark & - &0.4354&+4.5\%\\
\midrule
SimCast\textsubscript{6$\rightarrow$12} &\cmark&&0.4517&+8.4\%\\
SimCast\textsubscript{6$\rightarrow$12} &\cmark&\cmark&0.4521&+8.5\%\\
\midrule
SimCast\textsubscript{4$\rightarrow$12} &\cmark&&0.4472&+7.3\%\\
SimCast\textsubscript{4$\rightarrow$12} &\cmark&\cmark&0.4516& +8.4\%\\
\midrule
SimCast\textsubscript{1$\rightarrow$12} &\cmark&&0.4184 &+0.4\% \\
SimCast\textsubscript{1$\rightarrow$12} &\cmark&\cmark&0.4338 & +4.1\%\\
\bottomrule
\end{tabular}
\label{tab:model_justification}
\end{threeparttable}}
\vspace{-5mm}
\end{table}

\subsection{Ablation Studies}
\subsubsection{Model design justification} 
To evaluate the effectiveness of our proposed weighted MSE loss and short-to-long term knowledge distillation, we compared the performance obtained with and without each component and reported the results in Table~\ref{tab:model_justification}. SimCast\textsubscript{12} refers to a single SimCast model trained to simultaneously predict future 12 frames. A baseline CSI-M of 0.4166 was achieved, which improved to 0.4354 by applying our proposed weighted MSE loss. SimCast\textsubscript{6$\rightarrow$12} represents a nowcasting model sharing the same architecture as SimCast\textsubscript{12}. However, it was developed using our proposed two-step training strategy. First, a short-term model with $T_s^\prime=6$ was trained, followed by knowledge transfer to a long-term nowcasting model with $T_l^\prime=12$. We can see that, with autoregressive inference and knowledge distillation, CSI-M was successfully improved to 0.4517 and 0.4521, respectively.

\subsubsection{Impact of prediction horizon} As aforementioned, the prediction horizon is a key parameter that drives the model to learn and capture temporal dynamics within different ranges. Here we set $T_s^\prime$ to 1, 4, and 6, respectively, and compared the results in Table~\ref{tab:model_justification}. Generally speaking, a smaller $T_s^\prime$ allows the model to effectively capture short-term temporal patterns. However, its prediction accuracy may decline quickly when autoregressive inference is applied for longer lead times, as the model did not have a chance to learn long-term patterns. We observe that setting $T_s^\prime=6$ leads to the best performance improvement. Furthermore, our proposed short-to-long term knowledge distillation method enables the model to jointly learn from both short-term and long-term patterns, achieving greater stability compared to counterparts that do not incorporate this component.

\subsubsection{Impact of knowledge distillation} 
\begin{figure}[t]
    \centering
    \includegraphics[width=0.97\linewidth]{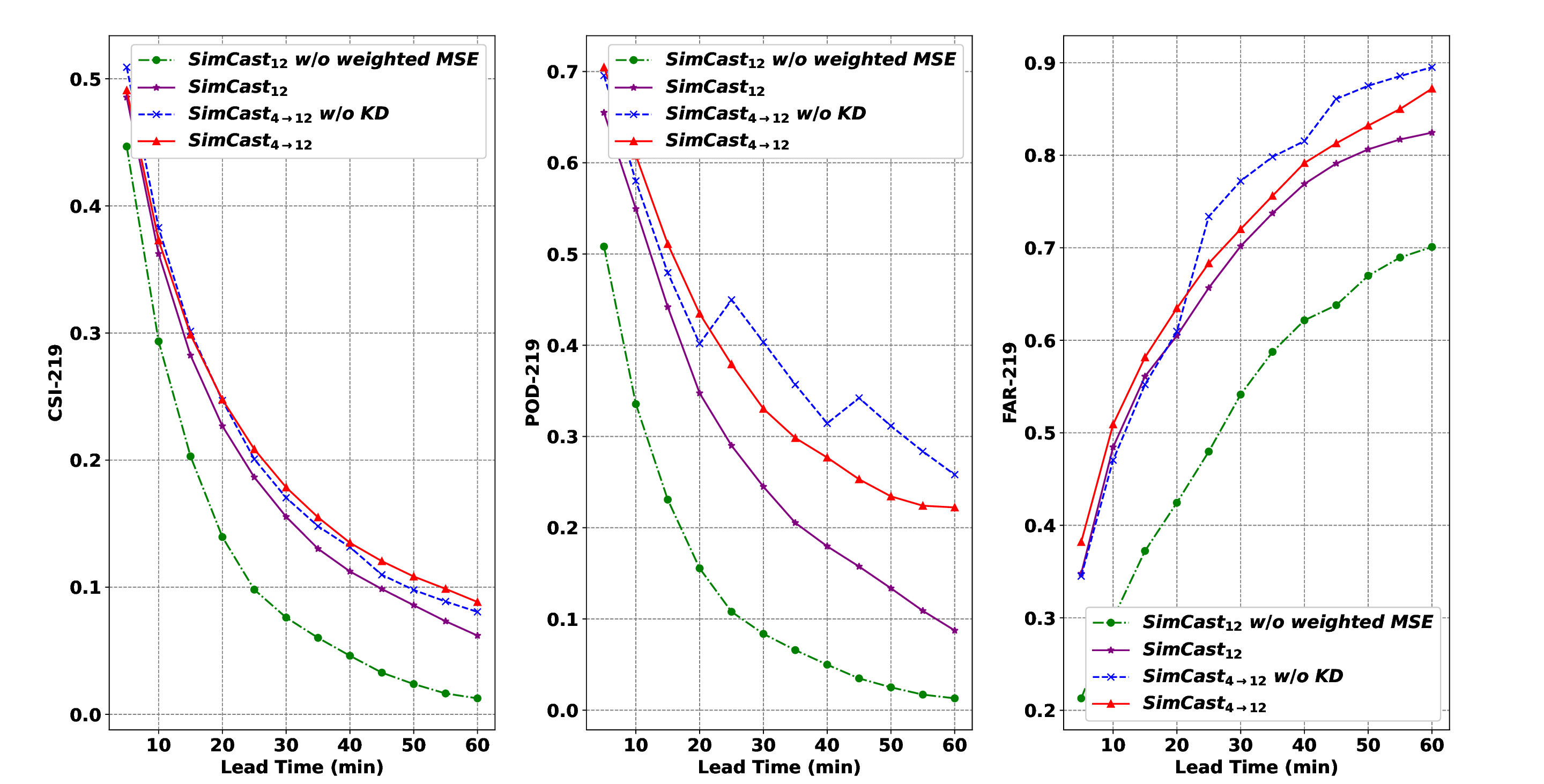}
    \caption{CSI, POD, FAR comparison at different lead time for threshold 219 on SEVIR dataset.}
    \label{fig:219}
    \vspace{-3mm}
\end{figure}

Fig.~\ref{fig:219} illustrates CSI, POD, and FAR comparison at different lead times for threshold 219 on the SEVIR dataset. POD and FAR stand for probability of detection and false alarm ratio, respectively. For method SimCast\textsubscript{4$\rightarrow$12} without knowledge distillation (blue dotted line), we can clearly observe a change in pattern every four frames caused by the autoregressive inference, especially for POD and FAR. Comparatively, the SimCast\textsubscript{12} method (purple solid line) exhibited smooth performance across varying lead times but achieved lower accuracy than SimCast\textsubscript{4$\rightarrow$12}. Method SimCast\textsubscript{4$\rightarrow$12} (red solid line) combines the strengths of learning from both short-term and long-term patterns, achieving the highest CSI scores for longer lead times and heavy rain rates, which are the most critical scenarios in radar-based nowcasting.

\subsubsection{Parameter sensitivity analysis}
Next, we study the impact of the pixel-wise weight $w_{max}$ in Eq.~\ref{eq:pixelwise_weight}. We set $w_{max}$ to different values, conduct experiments using method SimCast\textsubscript{12} on SEVIR dataset, and report the results in Table~\ref{tab:para}. Generally speaking, the mean CSI increases when a larger weight is adopted, with $w_{max}$ set to 10 by default in previous experiments. The results also indicate that our method remains robust across a wide range of $w_{max}$ values.

\begin{table}[t]
\centering
\caption{The effect of pixel-wise weight on SEVIR.}
    \resizebox{0.38\textwidth}{!}{
    \large
    \begin{tabular}{c| c c c c c }
        \toprule
       $w_{max}$ & 1 & 3&  5 & 7 &  10  \\
        \midrule
        CSI-M & 0.4166 & 0.4273 &0.4325& 0.4341 & 0.4354  \\
        \bottomrule
    \end{tabular}
    }
    \label{tab:para}
    \vspace{-5mm}
\end{table}

\begin{figure*}[t]
    \centering
    \includegraphics[width=0.85\textwidth]{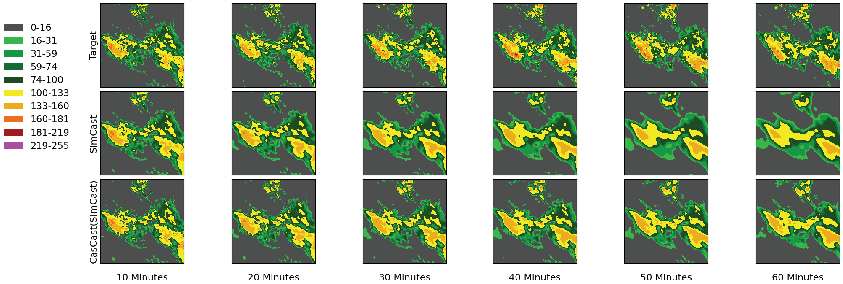}
    \caption{Qualitative comparison of the predictions generated by SimCast and CasCast(SimCast) at different lead times.}
    \label{fig:visualization}
\vspace{-4mm}
\end{figure*}
\subsubsection{Visualizations}
Fig.~\ref{fig:visualization} visualizes the ground-truth radar images and the prediction results generated by SimCast and CasCast(SimCast). Our proposed SimCast, being a deterministic model, is subject to common challenges associated with deterministic predictions, such as blurriness and distribution shift at longer lead times. Fortunately, the visual quality of deterministic predictions can be effectively improved by utilizing the capabilities of diffusion-based models, as demonstrated in recent work such as CasCast, NowcastNet, and DiffCast. These studies highlight that the performance of the deterministic model has a major impact on the results generated by the probabilistic model. Overall, our SimCast provides better deterministic predictions as conditional input, contributing to improved performance, as evidenced by the comparison between CasCast(SimCast) and CasCast(EarthFormer) in Table~\ref{tab:eval_sevir}.

\section{Conclusions}
We propose a simple yet effective precipitation nowcasting method named SimCast. Initially, a short-term nowcasting model is trained to serve as an auxiliary network, generating forecasted synthetic radar data to enrich the training samples. Coupled with a newly proposed knowledge distillation approach, a long-term nowcasting model with enhanced predictive capability is then trained using the augmented dataset. 
We evaluated our method using three public benchmark datasets, achieving state-of-the-art nowcasting performance, especially in regions of heavy rainfall.

\section{Acknowledgment}
This work is supported by the National Research Foundation, Prime Minister’s Office, Singapore under the Aviation Transformation Programme (Grant No. ATP2.0\_ATM-MET\_I2R). The authors would also like to thank our collaborators from Meteorological Service Singapore for providing valuable comments and suggestions on this work.
This work was also partially supported by the A*STAR Computational Resource Centre through the use of its high performance computing facilities.

\bibliographystyle{IEEEbib}
\bibliography{ref}

\end{document}